%% file: main.tex
\documentclass[10pt,twocolumn,letterpaper]{article}

\usepackage[pagenumbers]{cvpr}

\input{preamble}

\definecolor{cvprblue}{rgb}{0.21,0.49,0.74}
\usepackage[pagebackref,breaklinks,colorlinks,allcolors=cvprblue]{hyperref}

\def\confName{CVPR}
\def\confYear{2026}

\title{\raisebox{-0.25em}{\includegraphics[height=1.25em]{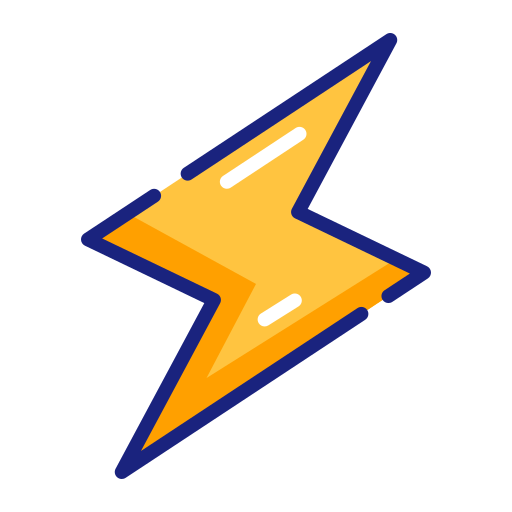}}LightSplat: Fast and Memory-Efficient Open-Vocabulary \\ 3D Scene Understanding in Five Seconds}

\author{
Jaehun Bang$^{1}$ \quad
Jinhyeok Kim$^{2}$\thanks{Work done while at UNIST.} \quad
Minji Kim$^{1}$ \quad
Seungheon Jeong$^{1}$ \quad
Kyungdon Joo$^{1}$\thanks{Corresponding author.} \\
$^{1}$ AIGS, UNIST \quad
$^{2}$ GSAI, POSTECH \quad \\
{\tt\small \{devappendcbangj, mzkim, sypsss, kyungdon\}@unist.ac.kr} \quad
{\tt\small jh4011@postech.ac.kr}
}

\begin{document}
\maketitle

% main
\input{sec/0_abs}
\input{sec/1_intro}
\input{sec/2_rw}
\input{sec/3_method}
\input{sec/4_exp}
\input{sec/5_conclusion}
\input{sec/6_ack}

{
    \small
    \bibliographystyle{ieeenat_fullname}
    \bibliography{main}
}

% supp
\clearpage
\appendix

\setcounter{figure}{0}
\setcounter{table}{0}
\setcounter{equation}{0}

\renewcommand{\thesection}{\Alph{section}}
\renewcommand{\thesubsection}{\thesection.\arabic{subsection}}
\renewcommand{\thefigure}{S\arabic{figure}}
\renewcommand{\thetable}{S\arabic{table}}
\renewcommand{\theequation}{S\arabic{equation}}

\twocolumn[{
\centering

{\Large \bf
\raisebox{-0.25em}{\includegraphics[height=1.2em]{fig/lightning.png}}
LightSplat: Fast and Memory-Efficient Open-Vocabulary \\
3D Scene Understanding in Five Seconds \\
Supplementary Materials
\par}

\vspace{15pt}
}]

\input{sec_supp/1_dataset}
\input{sec_supp/2_implementation}
\input{sec_supp/3_quantitative}
\input{sec_supp/4_qualitative}
\input{sec_supp/5_applications}
\input{sec_supp/6_limitations}

\begin{figure*}[t]
    \centering
    \includegraphics[width=0.9\linewidth]{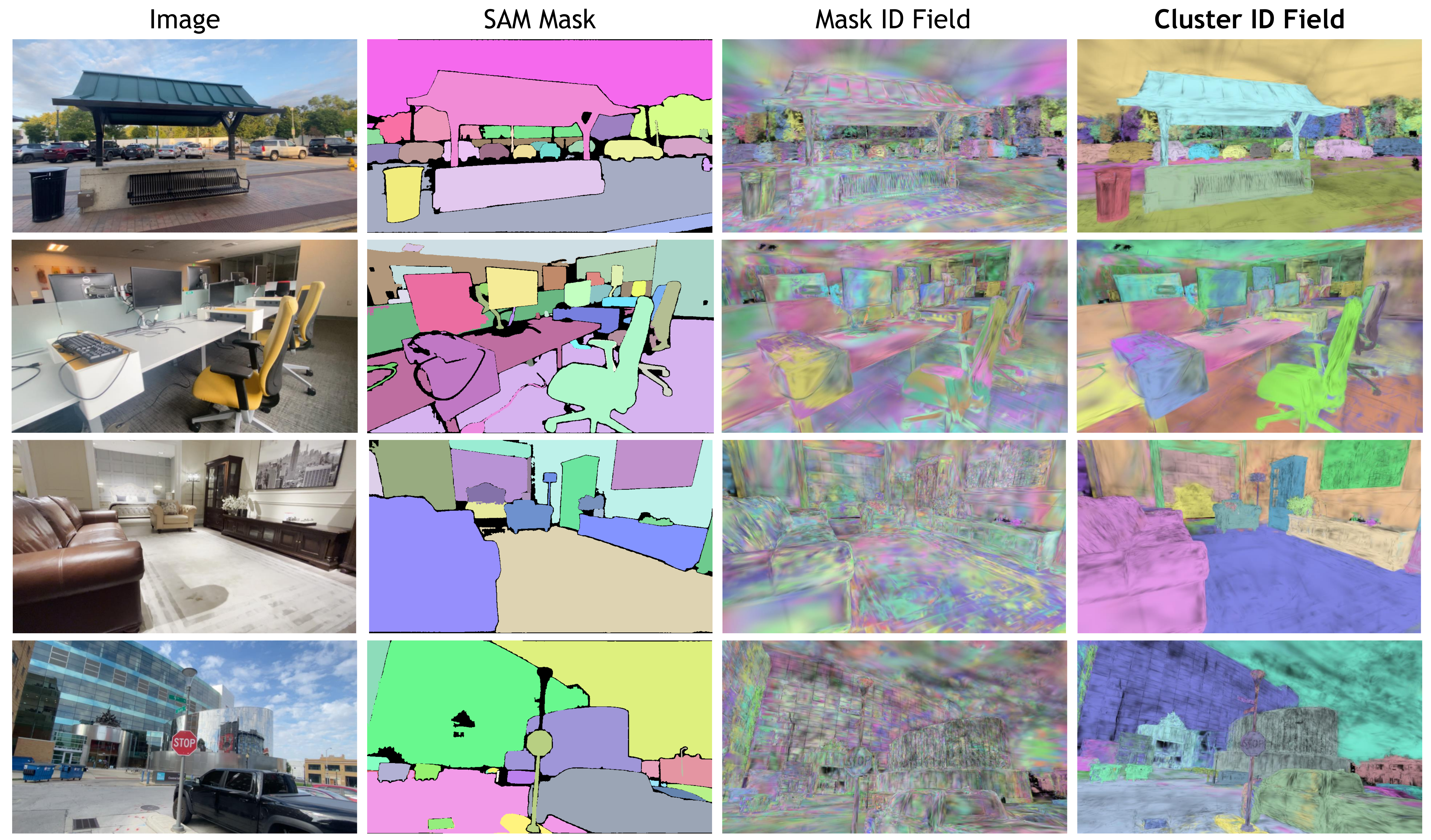}
    \caption{\textbf{Clustering Process Visualization.}
    The figure illustrates how SAM mask IDs are lifted into 3D and refined into coherent object-level clusters, resulting in the final cluster ID field.
    It demonstrates that our method produces clean cluster ID fields across diverse indoor and outdoor scenes without scene-specific tuning.
    }
    \label{fig:mask2cluster}
\end{figure*}

\begin{figure*}[t]
    \centering
    \includegraphics[width=1.0\linewidth]{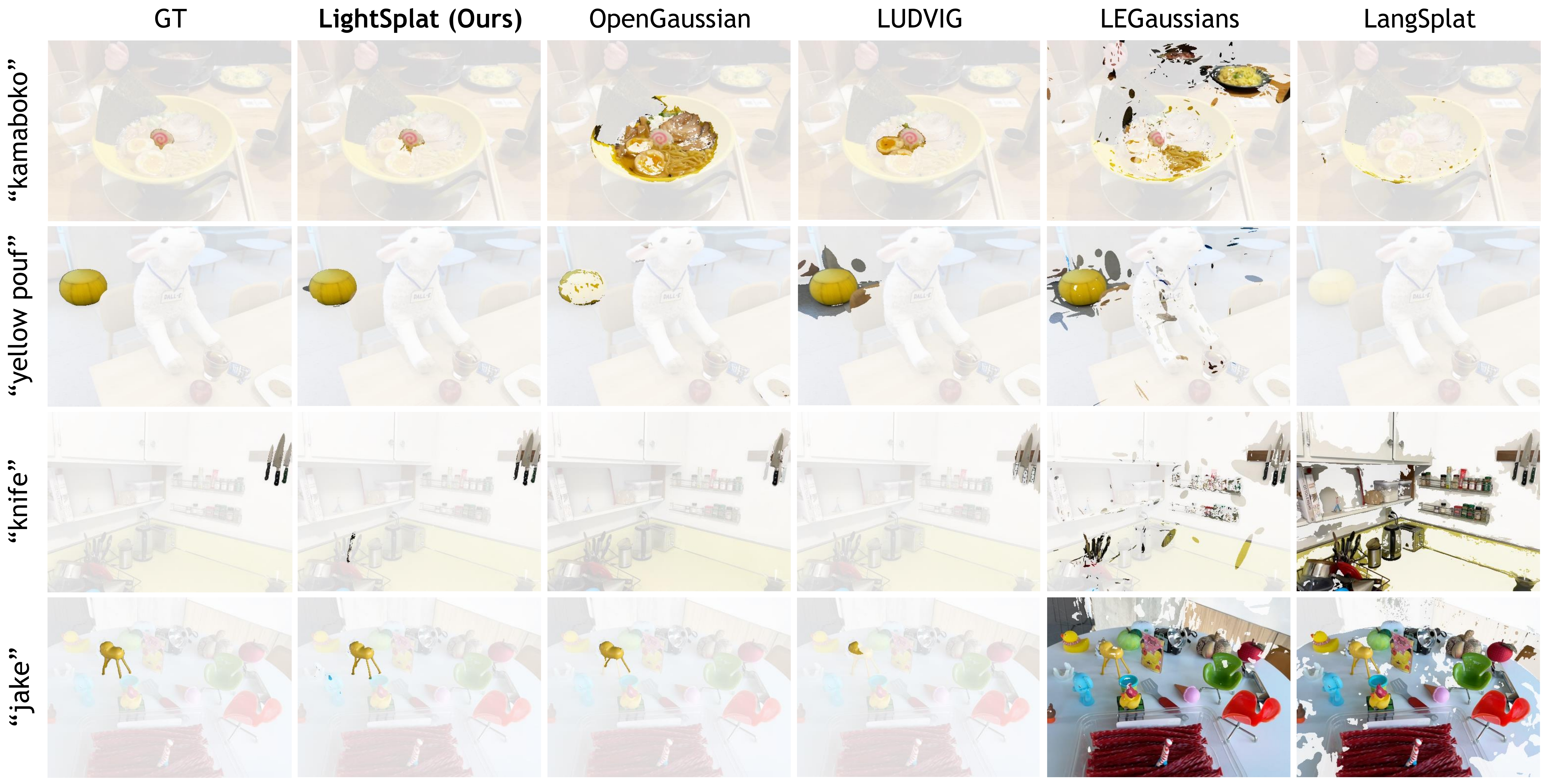}
    \caption{\textbf{More qualitative comparison for 3D Object Selection on the LERF-OVS dataset.}
    We visualize model performance across different scenes and text queries in LERF-OVS.
    With context-aware 3D clustering, our method delivers precise boundaries for challenging queries, including small objects in contact with others, repeated instances, and thin or intricate structures, while achieving the fastest performance.
    }
    \label{fig:more_qualitative_lerf}
\end{figure*}

\begin{figure*}[t]
    \centering
    \includegraphics[width=1.0\linewidth]{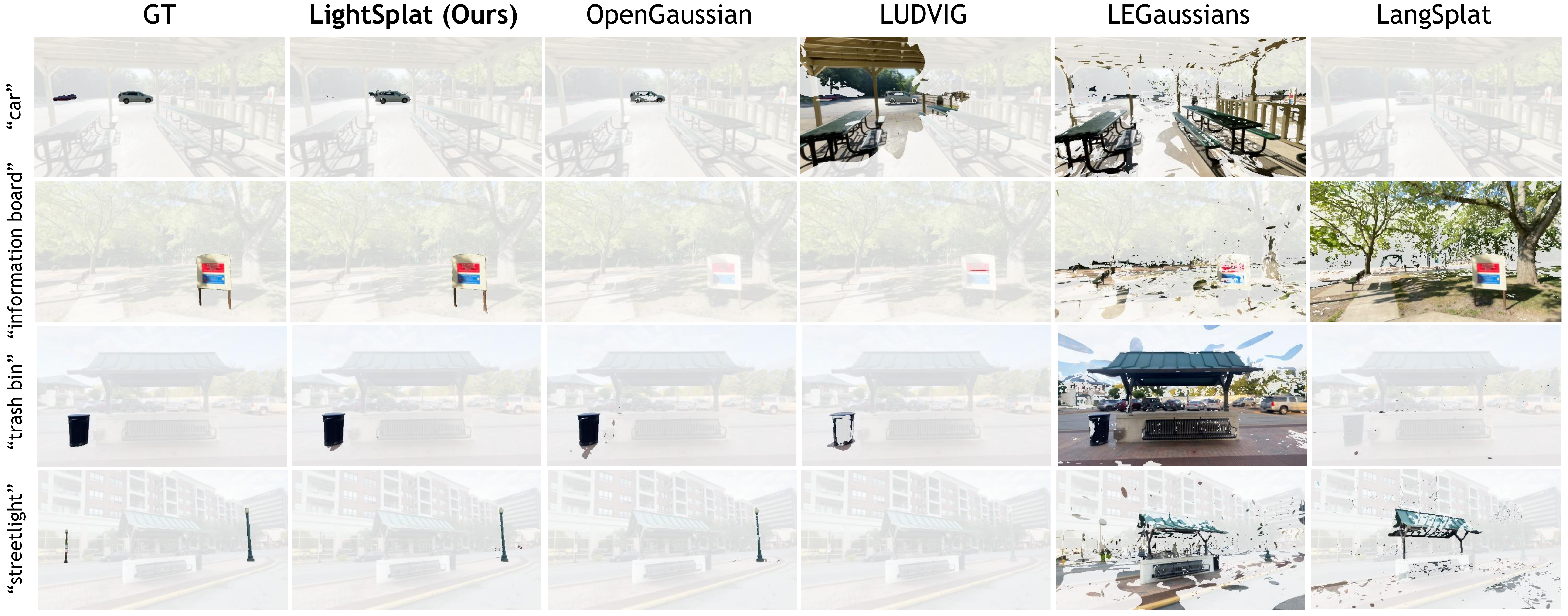}
    \caption{\textbf{More qualitative comparison for 3D Object Selection on the DL3DV-OVS dataset.}
    We visualize model performance across different scenes and text queries in DL3DV-OVS.
    With context-aware 3D clustering, our method produces accurate boundaries in large indoor-outdoor scenes, handling distant objects, multiple instances, and complex structures while maintaining the fastest performance.
    }
    \label{fig:more_qualitative_dl3dv}
\end{figure*}

\begin{figure*}[t]
    \centering
    \includegraphics[width=0.95\linewidth]{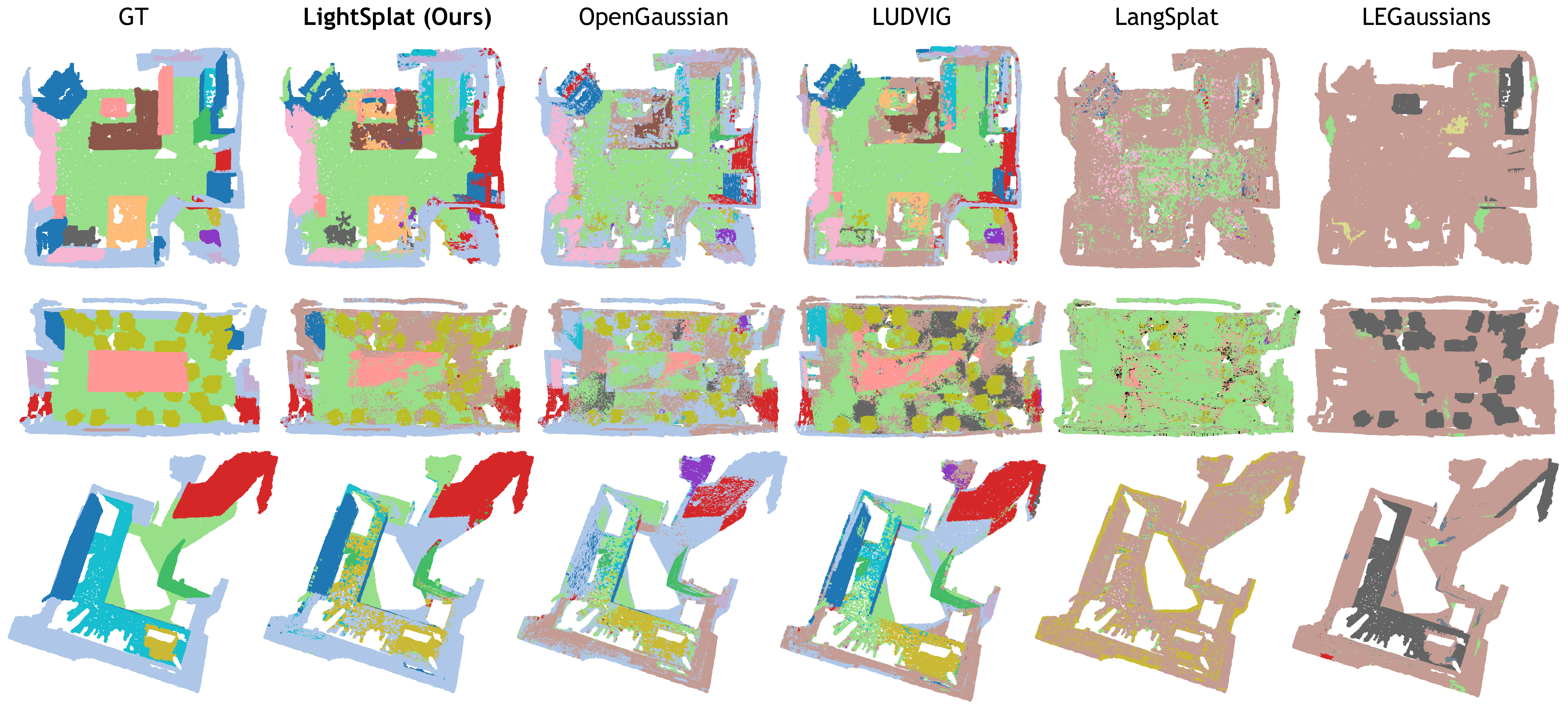}
    \caption{\textbf{More qualitative comparison for 3D Semantic Segmentation on the ScanNet dataset.}
    For intuitive comparison, ground-truth semantics are visualized in distinct colors.
    Our method captures a broader range of semantics in indoor scenes, providing more complete coverage with precise boundaries.
    It handles both object-level (e.g., door, cabinet) and large-area semantics (e.g., floor, wall), showing robustness across diverse scenes and outperforming other methods in accuracy and speed.
    }
    \label{fig:more_qualitative_scannet}
\end{figure*}

\begin{figure*}[t]
    \centering
    \includegraphics[width=0.95\linewidth]{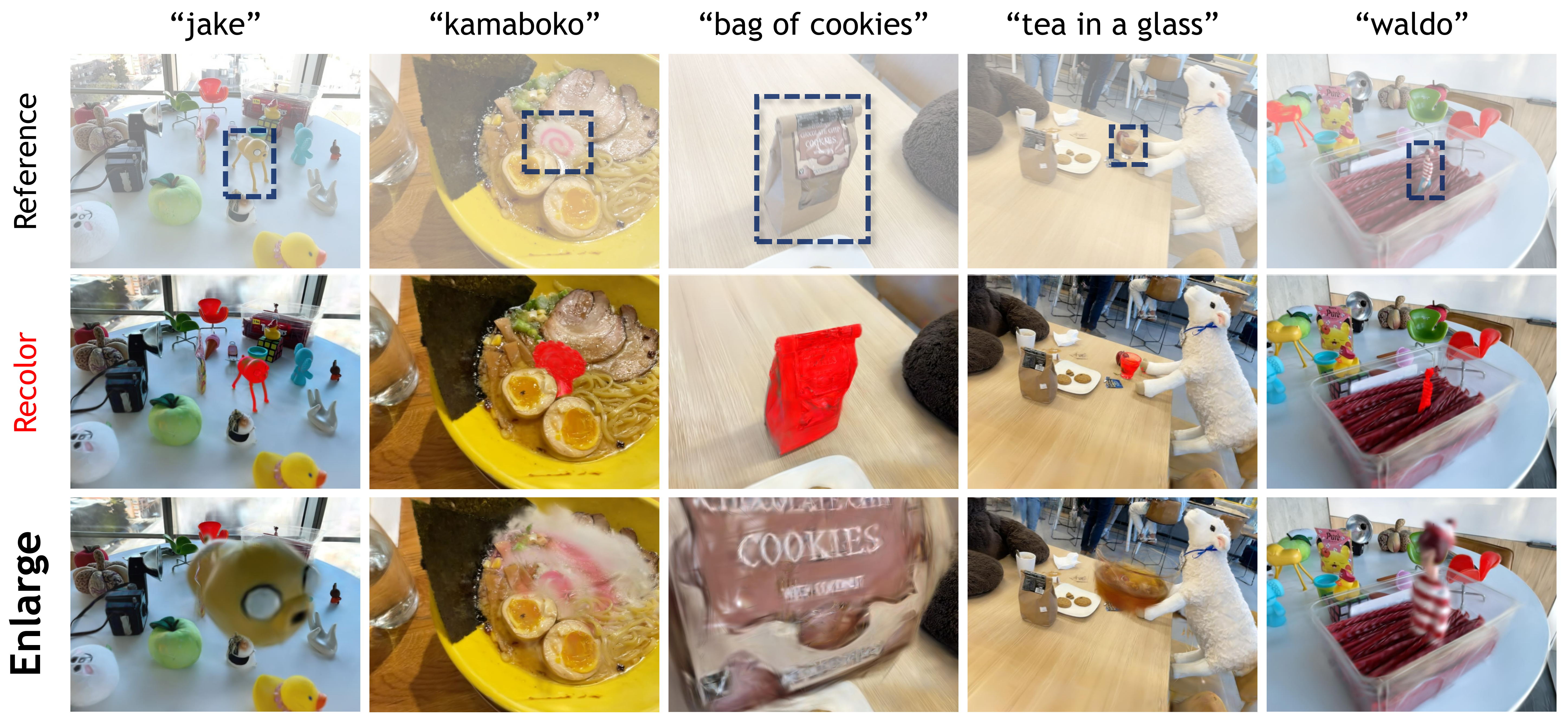}
    \caption{\textbf{Qualitative results of text-driven 3D scene editing on the LERF-OVS dataset.}
    Recoloring (middle) and enlarging (bottom) are shown as representative examples, enabled by our 3DGS-based approach for fast, accurate, and high-fidelity object manipulation across diverse scenes.
    }
    \label{fig:3d_editing}
\end{figure*}

\end{document}

%% file: preamble.tex
\usepackage{multirow}
\usepackage{booktabs}
\usepackage{makecell}
\usepackage[table]{xcolor}
\usepackage{colortbl}

\definecolor{tabred}{HTML}{F4BBBB}
\definecolor{taborange}{HTML}{FDD9A0}
\definecolor{tabred2}{HTML}{FF5A5A}

\newcommand{\best}{\cellcolor{tabred}}
\newcommand{\sbest}{\cellcolor{taborange}}

\newcommand{\Sref}[1]{Sec.~\ref{#1}}

\newcommand{\Fref}[1]{Fig.~\ref{#1}}
\newcommand{\Tref}[1]{Table~\ref{#1}}

\definecolor{brightcerulean}{rgb}{0.11, 0.67, 0.84}

%% file: sec/0_abs.tex
\begin{abstract}

Open-vocabulary 3D scene understanding enables users to segment novel objects in complex 3D environments through natural language.
However, existing approaches remain slow, memory-intensive, and overly complex due to iterative optimization and dense per-Gaussian feature assignments.
To address this, we propose \texttt{LightSplat}, a fast and memory-efficient training-free framework that injects compact 2-byte semantic indices into 3D representations from multi-view images.
By assigning semantic indices only to salient regions and managing them with a lightweight index-feature mapping, \texttt{LightSplat} eliminates costly feature optimization and storage overhead.
We further ensure semantic consistency and efficient inference via single-step clustering that links geometrically and semantically related masks in 3D.
We evaluate our method on LERF-OVS, ScanNet, and DL3DV-OVS across complex indoor-outdoor scenes.
As a result, \texttt{LightSplat} achieves state-of-the-art performance with up to 50-400$\times$ speedup and 64$\times$ lower memory, enabling scalable language-driven 3D understanding.
For more details, visit our project page {\small \url{https://vision3d-lab.github.io/lightsplat/}}.

\end{abstract}

%% file: sec/1_intro.tex
\section{Introduction}\label{sec:intro}

% intro
With growing demand for natural user interactions within 3D environments, open-vocabulary 3D scene understanding has emerged as an important task~\cite{kerr2023lerf, qin2024langsplat, bhalgat2024n2f2, shi2024legaussians, wu2024opengs, jun2025drsplat, marrie2025ludvig}.
It aims to identify and segment 3D scenes based on user queries, without predefined categories.
This flexibility allows handling complex environments with diverse and unpredictable objects, enabling real-world applications in robotic manipulation~\cite{ji2024graspsplats}, 3D scene editing~\cite{guo2024semantic}, and AR/VR.

% previous
A main challenge in this task is bridging the gap between language and 3D representations.
To address this, recent approaches distill 2D semantics into the 3D scene using object cues from SAM~\cite{kirillov2023sam} and semantic embeddings from CLIP~\cite{radford2021clip}.
As an early attempt, LERF~\cite{kerr2023lerf} embeds CLIP features into NeRF~\cite{mildenhall2021nerf}, but suffers from limited interactivity and scalability due to its implicit representation and high computational cost.
To enable efficient language-to-3D alignment, 3D Gaussian Splatting (3DGS)~\cite{kerbl20233dgs} provides an explicit representation with real-time rendering.
Leveraging this representation, recent approaches~\cite{qin2024langsplat, bhalgat2024n2f2, shi2024legaussians, wu2024opengs, jun2025drsplat, marrie2025ludvig} distill language semantics into the 3D scene.
These methods typically rely on rendering-guided iterative feature optimization, contrastive learning for semantic separation, and feature quantization to compress language features.

\begin{figure}[t]
    \centering
    \includegraphics[width=0.93\linewidth]{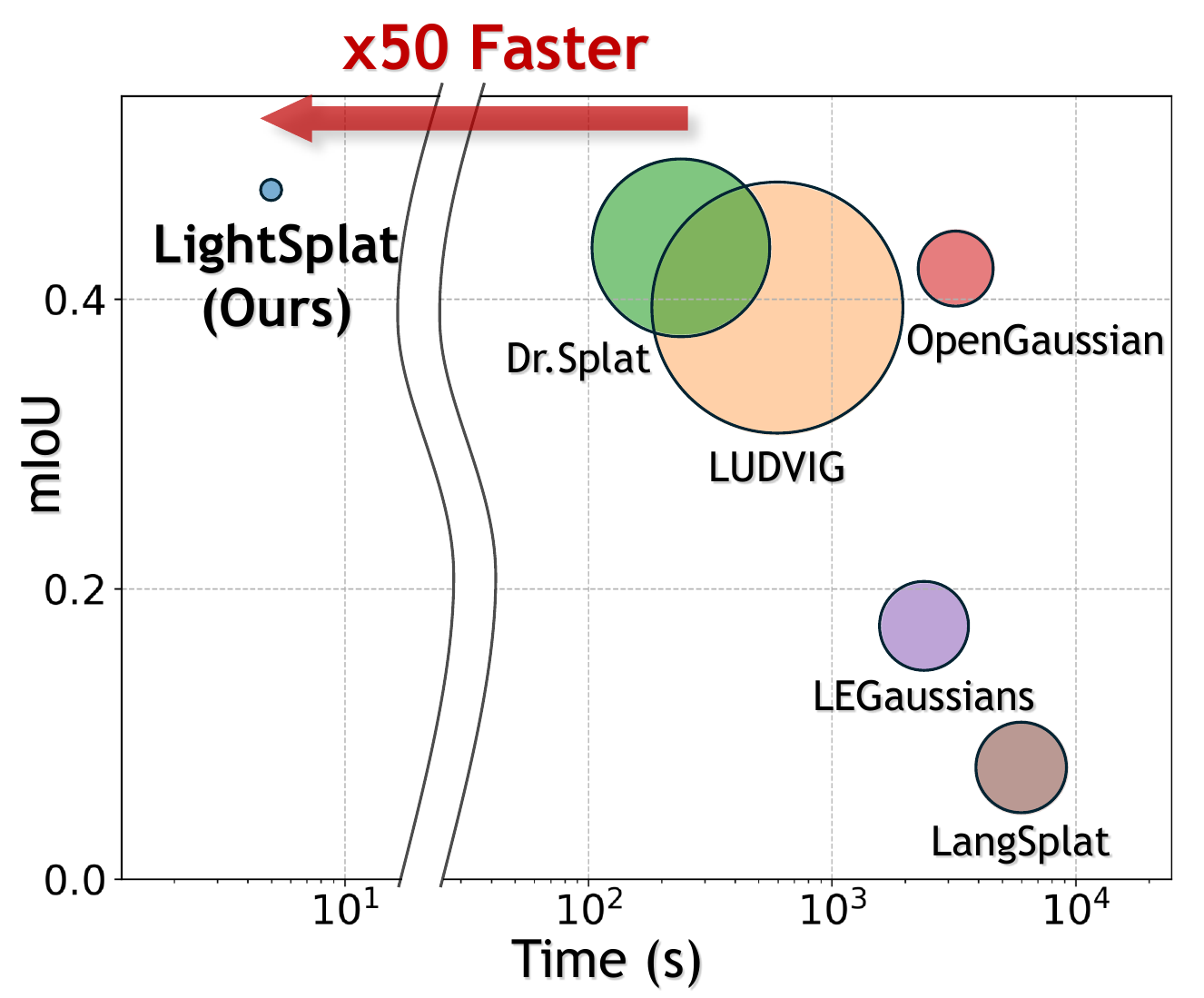}
    \caption{\textbf{Comprehensive comparison of speed, performance, and memory overhead.} 
    We evaluate recent open-vocabulary 3D scene understanding models in terms of distillation time (x-axis), segmentation performance (y-axis), and memory overhead (circle size).
    \texttt{LightSplat} achieves 50$\times$ faster feature distillation, higher accuracy, and 64$\times$ lower memory usage.
    LUDVIG’s circle is shown at half size because it is too large to display at full scale.
    }
    \label{fig:teaser}
\end{figure}

% limitations
Despite recent advances, existing methods still suffer from three major limitations: high computational cost, memory overhead, and semantic degradation, all of which hinder scalability in real-world scenarios.
First, feature distillation is bottlenecked by iterative optimization that repeatedly aligns rendered views with CLIP embeddings.
Although a direct mapping between 2D masks and 3D structure is feasible, this pipeline remains unnecessarily inefficient.
Second, assigning high-dimensional features to individual Gaussians leads to redundant storage and excessive per-Gaussian comparisons during inference.
Third, semantic degradation arises when Gaussians are projected into 2D, which blurs their features and leads to indirect supervision misaligned with 3D geometry.

% ours
We propose \texttt{LightSplat}, a fast and memory-efficient training-free framework for 3D scene understanding via lightweight semantic injection.
Our key insight is to bypass costly optimization by directly linking 2D semantics to 3D structure through discrete indexing.
In our method, we inject semantics only into Gaussians that have a high rendering contribution to the corresponding 2D masks.
Rather than storing full language features per Gaussian, we assign compact 2-byte mask indices, which are mapped to CLIP features via a lightweight mask-feature mapping.
This decouples feature storage from the 3D representation, eliminating per-Gaussian feature handling and significantly reducing memory and computational overhead.
To further improve inference speed and ensure semantic consistency in 3D, we cluster Gaussians based on geometric and semantic relationships, forming visually and conceptually meaningful objects.
By consolidating 2D mask features into 3D cluster representations using precomputed relationships among the 2D masks, we achieve fast and consistent object-level inference.
We evaluate \texttt{LightSplat} on LERF-OVS, ScanNet, and DL3DV-OVS, an open-vocabulary extension of DL3DV~\cite{ling2024dl3dv} for complex indoor-outdoor scenes.
With the streamlined design, \texttt{LightSplat} distills features in only 5 seconds, up to 50-400$\times$ faster than the previous state-of-the-art method~\cite{jun2025drsplat}, while outperforming it in both efficiency and performance as illustrated in \Fref{fig:teaser}.
Moreover, our method reduces memory usage by 64$\times$ with object-level indexed features instead of Gaussian-level language features.
This design provides a lightweight and scalable foundation for real-world scenarios~\cite{kerr2023lerf, dai2017scannet, ling2024dl3dv}.

In summary, our main contributions are as follows:
\begin{itemize}
    \item We propose \texttt{LightSplat}, a simple, training-free framework for open-vocabulary 3D scene understanding eliminating exhaustive iterative optimization.
    \item Our approach assigns each Gaussian a small index tied to object semantics via an index-feature mapping, enabling fast inference without per-Gaussian feature handling.
    \item We design a geometry and semantic-aware clustering method that constructs object-level representations in 3D, enabling efficient and interpretable inference.
    \item Our method distills semantics in 5 seconds, achieving 50-400$\times$ speedup and 64$\times$ lower memory than prior SOTA, while preserving high-quality 3D semantics.
\end{itemize}

%% file: sec/2_rw.tex
\section{Related Work}\label{sec:rw}

\begin{figure*}[t]
    \centering
    \includegraphics[width=0.87\linewidth]{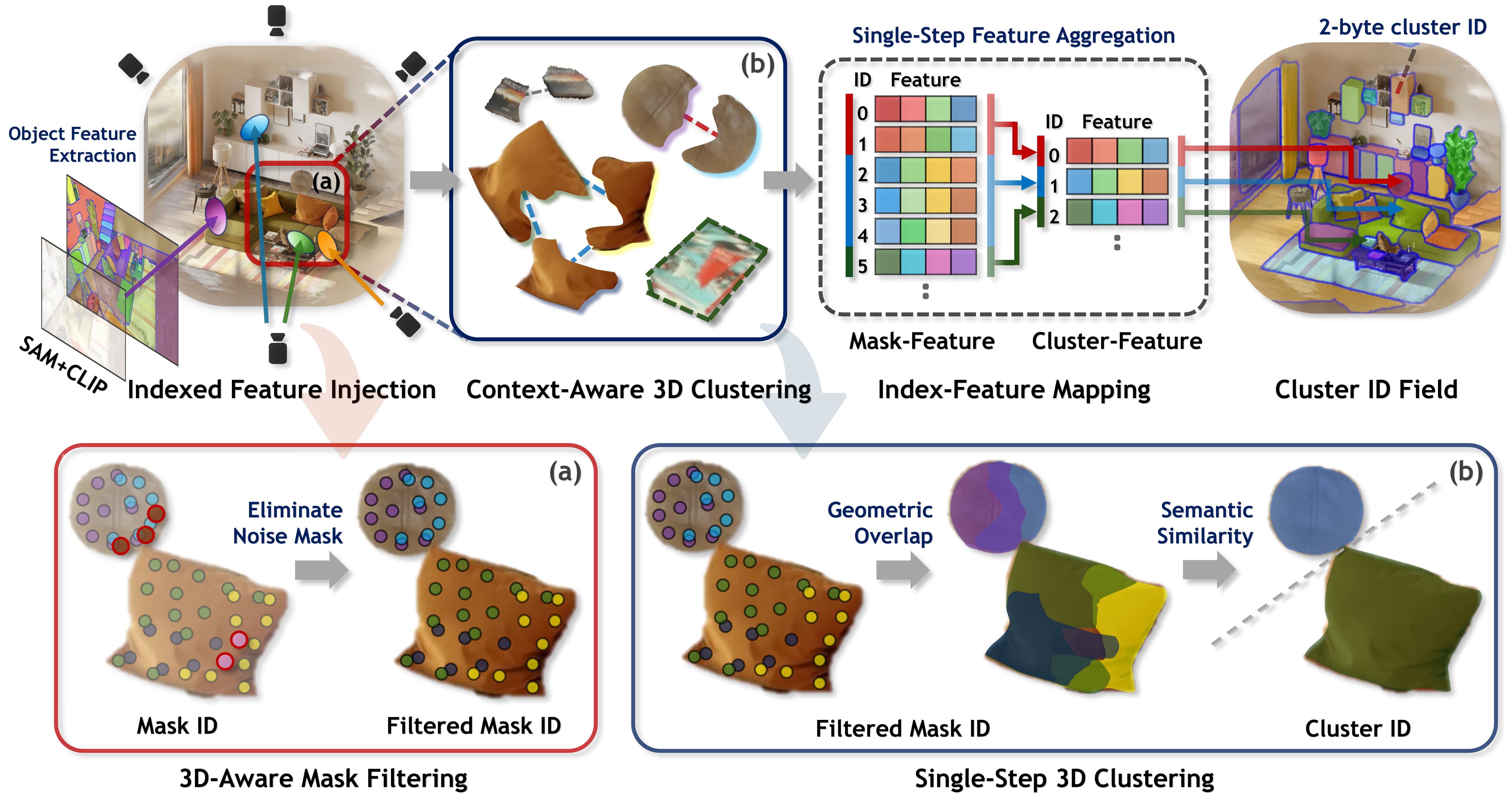}
    \vspace{-2mm}
    \caption{\textbf{Overall framework of \texttt{LightSplat}.}
    From multi-view images, we obtain SAM masks and corresponding CLIP features.
    We align them to the 3D scene via indexed feature injection, assigning each Gaussian a compact 2-byte mask index based on its projection influence.
    (a) To ensure semantic consistency, we perform 3D-aware mask filtering, and (b) construct an inter-mask graph via index-feature mapping based on geometric and semantic relations, which guides context-aware 3D clustering in a single step.
    This enables cluster-level feature management with a compact cluster ID field for efficient, interpretable, and training-free open-vocabulary 3D scene understanding.
    }
    \vspace{-2mm}
    \label{fig:framework}
\end{figure*}

\paragraph{3D Scene Representations.}
3D scene representations reconstruct 3D environments from multi-view images, facilitating novel view synthesis and 3D scene understanding.
NeRF provides high-quality rendering via MLP-based volumetric fields, but remains slow and inefficient for real-world applications despite efforts to accelerate it~\cite{neff2021donerf, garbin2021fastnerf, xu2022point, yan2023plenvdb, wang2023f2}.
To overcome these computational bottlenecks, 3DGS has emerged as a promising alternative, offering real-time rendering and explicit scene representation.
It models a scene using a collection of 3D Gaussian primitives, which can be efficiently rendered through differentiable rasterization.
As an explicit representation, 3DGS offers higher interpretability and direct access to its primitives, enabling easier integration into various applications~\cite{chen2024gaussianeditor, jiang2024vr, yu2024cogs, wu2024gaussctrl}.

In this work, we leverage the real-time and explicit nature of 3DGS to introduce an efficient and straightforward method for injecting semantics into 3D scenes.

\paragraph{Open-Vocabulary 3D Scene Understanding.}
Given a natural language query, the goal of open-vocabulary 3D scene understanding is to segment corresponding objects in a 3D scene.
Without relying on predefined categories, this approach enables more flexible and scalable scene understanding in real-world environments, where object categories are diverse.
Many studies have explored integrating language features into 3D scene representation, focusing on distilling semantic embeddings from 2D foundation models like CLIP~\cite{radford2021clip}, DINO~\cite{caron2021emerging}, and SAM~\cite{kirillov2023sam} to the radiance field.

Early works~\cite{kobayashi2022decomposing, tschernezki2022neural, kerr2023lerf, tsagkas2023vl} introduce additional networks into NeRF to learn semantic features.
DFF~\cite{kobayashi2022decomposing} augments the network for feature prediction, and LERF~\cite{kerr2023lerf} uses multi-scale CLIP features as supervision.
While these methods offer a unified 3D-language representation, they are limited by NeRF’s slow and computationally intensive volumetric rendering.

To address this, recent studies~\cite{qin2024langsplat,bhalgat2024n2f2,shi2024legaussians,liang2024supergseg,zhang2025bootstraping} adopt 3DGS and optimize language features through rendering-based iterative training.
This optimization gradually enriches the 3D Gaussians with language semantics, leading to interpretable 3D representations.
For example, LangSplat~\cite{qin2024langsplat} and LEGaussians~\cite{shi2024legaussians} distill CLIP features and compress them using an autoencoder and codebooks.
While these methods provide a useful representation, they suffer from slow feature distillation caused by iterative feature optimization.
Moreover, because the segmentation happens on 2D rendered feature maps, the model cannot directly localize the corresponding objects in 3D space.
OpenGaussian~\cite{wu2024opengs} performs segmentation in 3D space, but still relies on an iterative process to train features.
Furthermore, its codebook-based feature quantization struggles to fully cover the scene when the scene complexity exceeds the codebook's capacity.

Recent works such as Dr.Splat~\cite{jun2025drsplat} and LUDVIG~\cite{marrie2025ludvig} associate CLIP features directly with 3D Gaussians by projecting 2D features onto them.
However, both still rely on iterative feature aggregation and store high-dimensional features for every Gaussian, leading to unnecessary computational and memory cost.
In addition, Dr.Splat requires large-scale Product Quantization training, while LUDVIG employs graph diffusion, further increasing computational overhead.
These limitations highlight the need for a simpler semantic representation that stores information only where it is useful, reducing both memory use and computation.
Motivated by these limitations, we propose a training-free and memory-efficient method that represents object-level semantics using compact indices, which are linked to features via a lightweight index-feature mapping.

%% file: sec/3_method.tex
\section{Method}\label{sec:method}

\subsection{Overview}\label{sec:3.1}
The key insight of \texttt{LightSplat} is that object semantics in 2D masks can be directly lifted to 3D via compact mask indices.
This enables single-step semantic injection and inter-mask clustering without per-Gaussian features.
As a result, it supports fast 2D-3D semantic alignment and scalable deployment without iterative training.

\begin{figure}[t]
    \centering
    \includegraphics[width=1.0\linewidth]{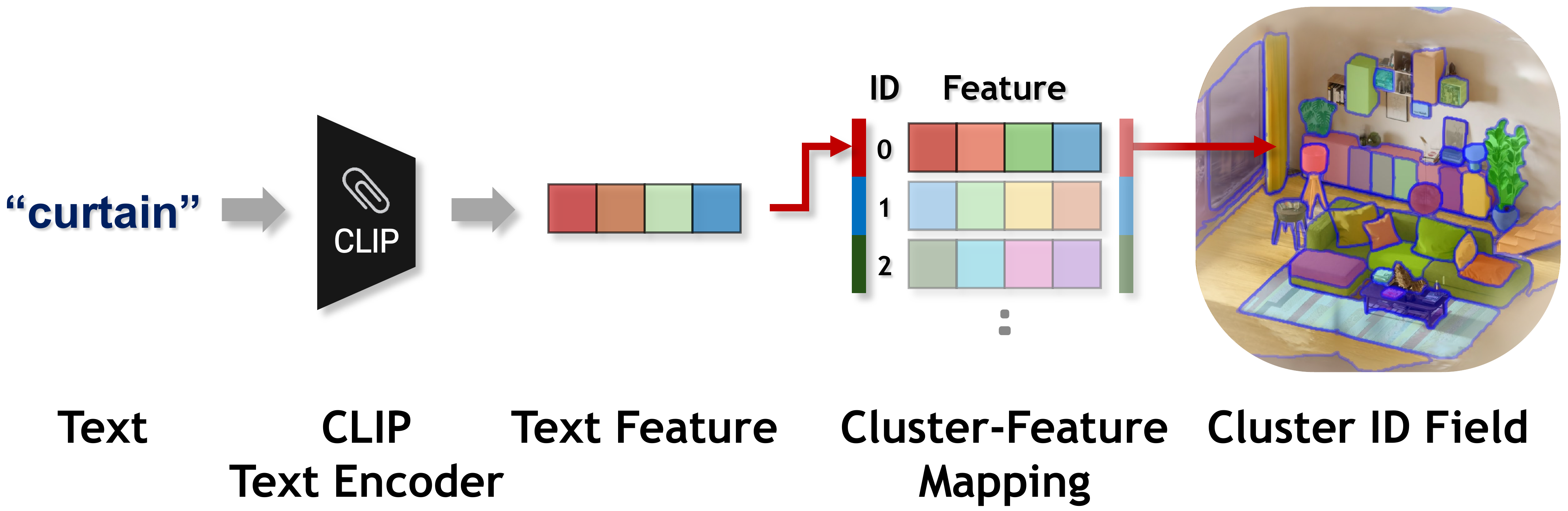}
    \caption{\textbf{Fast inference via cluster-feature mapping.}
        During inference, the text query is compared with a compact set of cluster features instead of all Gaussians or pixels, enabling fast retrieval.
    }
    \label{fig:inference}
\end{figure}

\Fref{fig:framework} illustrates the overall \texttt{LightSplat} framework.
The pipeline begins by extracting 2D object masks and their corresponding CLIP features from multi-view images using SAM and CLIP, following prior work~\cite{qin2024langsplat}.
Instead of storing high-dimensional features per Gaussian, we assign compact 2-byte mask indices based on their contribution to the 2D mask.
To improve the reliability of this mapping, we filter out noisy masks that have limited impact on the 3D scene.
This reduces view-dependent artifacts and strengthens multi-view semantic consistency.
To manage semantics efficiently, we propose an index-feature mapping that associates each 2-byte index to its corresponding CLIP feature.
This design replaces redundant per-Gaussian features with a compact object-level representation, allowing fast and memory-efficient inference.
Finally, we perform context-aware 3D clustering to group Gaussians into object-level clusters.
To achieve this, we construct a graph over the filtered 2D masks based on their geometric overlap in 3D and semantic similarity, and partition it in a single step.
We then assign each 3D cluster a representative language feature, enabling compact and interpretable object-level inference, as illustrated in ~\Fref{fig:inference}.

\subsection{Index-Feature Mapping}\label{sec:3.2}
To overcome the inefficiency of iterative per-Gaussian optimization, we replace mask-level semantic features with compact 2-byte indices.
These indices are linked to CLIP features via a mask-feature mapping, allowing each Gaussian to store only a mask index instead of high-dimensional language features.
Each mask index acts as a key to its language feature in the index-feature mapping tensor.
During clustering, we consolidate the mask-feature mapping into a compact cluster-feature mapping by grouping related masks across multiple views in a single step.
These mask-feature and cluster-feature mappings collectively form the index-feature mapping.
This index-feature mapping is not merely a storage reduction technique.
It serves as the core design that accelerates the pipeline, preserves semantic consistency across views, and enables scalable 3D understanding.

\subsection{Indexed Feature Injection}\label{sec:3.3}
This section describes a single-step transfer of 2D semantics into 3D while preserving semantic information and removing iterative optimization.
To achieve this, we exploit the mask-feature mapping by decoupling feature storage from the 3D representation and injecting compact mask indices into the 3D scene.

\paragraph{Object Feature Extraction.}
We first extract object semantics from each view, which serves as a crucial foundation for composing 3D object semantics.
Specifically, given multi-view images $\textbf{I} = \{ {I_l} \}_{l=1}^L$ where $L$ is the total number of frames, each image $I_l \in \mathbb{R}^{3 \times H \times W}$ is of size $H \times W$, with 3 color channels.
For each view, we use SAM to extract object masks $\mathcal{M}$:
\begin{equation}
\mathcal{M} = \{ m_k \}_{k=1}^{K}, \quad m_k \in \{ 0,1 \}^{H\times W},
\end{equation}
where $K$ is the total number of masks.
For each mask, we extract the corresponding CLIP features:
\begin{equation}
\mathcal{F} = \{ f_k \}_{k=1}^{K}, \quad f_k \in \mathbb{R}^{512}.
\end{equation}
We then assign each 2D mask a unique index to manage its CLIP features and inject semantics efficiently into 3DGS.

\noindent \textbf{Gaussian Contribution.}
To assign semantics only to Gaussians that significantly contribute to the rendered image, we compute their pixel-wise contributions using alpha-blending weights from the rendering equation~\cite{kerbl20233dgs}.
The $n$-th Gaussian contribution at pixel $(u,v)$ in the $l$-th view is defined as:
\begin{equation}
w_n^{(l)}(u,v) = \alpha_n \cdot T_n^{(l)}(u,v),
\end{equation}
where $\alpha_n$ is the opacity of the $n$-th Gaussian.
The transmittance \( T_n^{(l)}(u,v) \) quantifies how much light reaches the $n$-th Gaussian at pixel \((u,v)\) in the \(l\)-th view.
It is computed by accumulating the occlusion effects from all preceding Gaussians along the same ray:
\begin{equation}
T_n^{(l)}(u,v) = \prod_{j=1}^{n-1} \left(1 - \alpha_j(u,v)\right),
\end{equation}
where \( \alpha_j(u,v) \) denotes the opacity of the \(j\)-th Gaussian at that pixel.

\paragraph{Indexed Feature Injection.}
To achieve efficient semantic injection, we assign 2-byte mask indices instead of full language features to Gaussians that contribute meaningfully in the image space:
\begin{equation}
\mathcal{G}_k = \left\{ g_n \;\middle|\; m_k^{(l)}(u,v)=1,\; w_n^{(l)}(u,v) \ge \tau_{\text{contrib}} \right\},
\end{equation}
where $\mathcal{G}_k$ is the set of Gaussians $g_n$ associated with mask $m_k$, with $\tau_{\text{contrib}}$ for contribution filtering.
We assign semantics only to Gaussians that make a meaningful visual contribution, avoiding fixed-size selection (\emph{e.g.}, top-k) that can include Gaussians with negligible visual impact.
Our approach significantly reduces memory usage by 1024 times compared to storing CLIP features directly, reducing the size from 4$\times$512 bytes to just 2 bytes per Gaussian.
This enables lightweight mask-level feature management and scalable object-level reasoning in 3D clustering without accessing per-Gaussian features.

\paragraph{3D-Aware Mask Filtering.}
To enhance semantic reliability and efficiency, we filter out masks that contribute insufficiently to 3D feature construction:
\begin{equation}
{\mathcal{M}_{\text{filtered}} = \left\{ m_k \;\middle|\; |\mathcal{G}_k| \ge \tau_{noise} \right\}},
\end{equation}
where $|\mathcal{G}_k|$ denotes the number of Gaussians associated with the $k$-th SAM mask, measuring its geometric support in 3D.
This step leverages Gaussian contributions from 2D-3D correspondences to suppress view-dependent artifacts while preserving masks with sufficient geometric support under $\tau_{\text{noise}}$.
Finally, we assign each Gaussian the most influential mask index across multiple views.

As a result, indexed feature injection compactly transfers 2D semantics into 3D based on each Gaussian contribution, forming the basis for scalable object-level reasoning.

\subsection{Context-Aware 3D Clustering}\label{sec:3.4}
To achieve more scalable and interpretable 3D understanding, we cluster Gaussians into object-level groups using semantic and spatial context.
Leveraging the mask indices from the previous stage, our method first connects semantically related 2D masks across views.
Their associated Gaussians are then grouped into coherent 3D clusters without storing high-dimensional per-Gaussian features.
This compact representation efficiently transfers 2D mask semantics into 3D coherent objects, enabling faster, memory-efficient, and consistent multi-view understanding.

\paragraph{Single-Step 3D Clustering.}
To enable object-level 3D clustering in a single step, we construct an undirected graph $G$, where each node $V$ represents a filtered 2D mask $m_k \in \mathcal{M}_{\mathrm{filtered}}$, and edges $E$ represent connectivity between these masks:
\begin{equation}
G = (V, E),\quad V = \left\{ \nu \;\middle|\; m_k \in \mathcal{M}_{\mathrm{filtered}} \right\}.
\end{equation}
We initialize $E$ with self-loop edges, defining \( E = \{ (k, k) \mid k \in V \} \), so that each node initially forms its own cluster.
To determine mask connectivity in the graph, we measure their geometric overlap and semantic similarity between masks $m_k$ and $m_{k'}$ in 3D space.
First, we compute the geometric overlap using the Intersection over Union (IoU) of their corresponding Gaussian sets:
\begin{equation}
\operatorname{IoU}\bigl(\mathcal{G}_k, \mathcal{G}_{k'}\bigr) = \frac{|\mathcal{G}_k \cap \mathcal{G}_{k'}|}{|\mathcal{G}_k \cup \mathcal{G}_{k'}|}.
\end{equation}
Second, we compute the semantic similarity using the cosine similarity between their CLIP features:
\begin{equation}
\text{sim}(f_k, f_{k'}) = \frac{f_k \cdot f_{k'}}{\|f_k\| \, \|f_{k'}\|}.
\end{equation}
Then, we connect masks $k$ and $k'$ with an edge if their geometric and semantic agreement suggests that they correspond to the same object:
\begin{equation}
(k, k') \in E\quad \text{if} \quad\begin{cases}\operatorname{IoU}(\mathcal{G}_k, \mathcal{G}_{k'}) \ge \tau_{\mathrm{IoU}} \\\mathrm{sim}(f_k, f_{k'}) \ge \tau_{\mathrm{feat}}\end{cases},
\end{equation}
with $\tau_{\mathrm{IoU}}$ and $\tau_{\mathrm{feat}}$ as cutoffs for geometric overlap and semantic similarity, respectively.
Each cluster corresponds to a connected component of $G$.
\begin{equation}
(k, k') \in E \quad \implies \quad \mathcal{C}_k = \mathcal{C}_{k'}.
\end{equation}

\paragraph{Single-Step Feature Aggregation.}
For efficient feature management, we compute the cluster-level semantic features in a single step by directly averaging the CLIP features of all associated masks.
Since mask-feature mappings are already established and clustering operates at the mask-level, the cluster-feature mapping can be constructed immediately in this single step.
For example, inference complexity can be reduced from comparing over 100,000 Gaussians to only about 100 clusters, lowering similarity computations to below 0.1\% depending on scene complexity.

Overall, our clustering strategy provides view-consistent 3D semantics while enabling fast distillation, efficient inference, and compact memory usage through jointly leveraging geometric and semantic cues.

%% file: sec/4_exp.tex
\section{Experiments}\label{sec:exp}

\subsection{Experimental Setup}\label{sec:4.1}

\begin{table*}[!ht]
    \centering
    \caption{\textbf{Quantitative comparison for 3D Object Selection on the LERF-OVS dataset.} 
    \textcolor{tabred2}{Red} and \textcolor{orange}{orange} highlight the best and second-best results in each column.
    FD Time refers to feature distillation time. If FD Time exceeds 20 minutes, it is rounded at the ones place.
    }
    \resizebox{0.90\linewidth}{!}{ 
        \begin{tabular}{c|ccccc|ccccc|c}
            \toprule
            \multirow{2}{*}{Methods} & \multicolumn{5}{c|}{mIoU} & \multicolumn{5}{c|}{mAcc @ 0.25} & \multirow{2}{*}{\textbf{FD Time}} \\
            & Waldo & Ramen & Figurines & Teatime & \textbf{Mean}
            & Waldo & Ramen & Figurines & Teatime & \textbf{Mean} \\
            \midrule
            LangSplat & 9.61 & 3.49 & 9.64 & 7.91 & 7.66
            & 9.09 & 5.63 & 14.29 & 8.47 & 9.37 & 100 min \\
            LEGaussians & 17.21 & 14.15 & 16.40 & 21.93 & 17.42
            & 27.27 & 28.17 & 25.00 & 33.90 & 28.56 & 40 min \\
            OpenGaussian & 30.96 & 24.02 & \best 55.38 & \sbest 58.24 & 42.15
            & 45.45 & 28.17 & 75.00 & 76.27 & 56.22 & 50 min \\
            LUDVIG & 28.08 & \sbest 29.33 & 47.43 & 52.28 & 39.28
            & 51.33 & \sbest 50.36 & 67.02 & \best 85.72 & 63.61 & 10 min \\
            Dr.Splat & \best 39.07 & 24.70 & \sbest 53.36 & 57.20 & \sbest 43.58
            & \best 63.64 & 35.21 & \best 80.36 & 76.27 & \sbest 63.87 & \sbest 4 min \\
            Ours & \sbest 34.95 & \best 45.07 & 50.63 &  \best 59.65 &  \best \textbf{47.58}
            & \sbest 59.09 & \best 57.75 & \sbest 76.79 & \sbest 79.66 & \best \textbf{68.32} & \best \textbf{4.2 s} \\
            \bottomrule
        \end{tabular}
    }
    \label{tab:quantitative_lerf}
\end{table*}

\begin{table*}[!ht]
    \centering
    \caption{\textbf{Quantitative comparison for 3D Object Selection on the DL3DV-OVS dataset.}
    \textcolor{tabred2}{Red} and \textcolor{orange}{orange} highlight the best and second-best results in each column.
    FD Time refers to feature distillation time. If FD Time exceeds 20 minutes, it is rounded at the ones place.
    }
    \resizebox{0.90\linewidth}{!}{ 
        \begin{tabular}{c|ccccc|ccccc|c}
            \toprule
            \multirow{2}{*}{Methods} & \multicolumn{5}{c|}{mIoU} & \multicolumn{5}{c|}{mAcc @ 0.25} & \multirow{2}{*}{\textbf{FD Time}} \\
            & Park & Shop & Road & Office & \textbf{Mean}
            & Park & Shop & Road & Office & \textbf{Mean} \\
            \midrule
            LangSplat & 8.79 & 9.28 & 7.90 & 15.80 & 10.44
            & 0.00 & 17.65 & 16.67 & 27.27 & 15.40 & 220 min \\
            LEGaussians & 21.95 & 9.70 & 3.51 & 8.17 & 10.83
            & 33.33 & 11.76 & 0.00 & 0.00 & 11.27 & 40 min \\
            OpenGaussian & 29.65 & 19.51 & \best 41.59 & 6.78 & 24.38
            & 41.67 & 35.29 & \best 66.67 & 0.00 & 35.91 & 60 min \\
            LUDVIG & \sbest 37.03 & \sbest 33.56 & 27.68 & \sbest 18.58 & \sbest 29.21
            & \sbest 70.00 & \best 56.75 & \sbest 53.57 & \sbest 47.22 & \sbest 56.89 & \sbest 12 min \\
            Ours & \best 69.78 & \best 35.59 & \sbest 31.43 & \best 43.13 & \best \textbf{44.98}
            & \best 91.67 & \sbest 47.06 & 50.00 & \best 54.55 & \best \textbf{60.82} & \best \textbf{4.8 s} \\
            \bottomrule
        \end{tabular}
    }
    \label{tab:quantitative_dl3dv}
\end{table*}

\begin{table*}[!ht]
    \centering
    \caption{\textbf{Quantitative comparison for 3D Semantic Segmentation on the ScanNet dataset.} 
    \textcolor{tabred2}{Red} and \textcolor{orange}{orange} highlight the best and second-best results in each column.
    FD Time, Runtime, and Memory indicate the feature distillation time, average inference time per text query, and feature size per Gaussian, respectively. If FD Time exceeds 20 minutes, it is rounded at the ones place.
    }
    \resizebox{0.86\linewidth}{!}{ 
        \begin{tabular}{c|cc|cc|cc|ccc}
            \toprule
            \multirow{2}{*}{Methods} & \multicolumn{2}{c|}{19 classes} & \multicolumn{2}{c|}{15 classes} & \multicolumn{2}{c|}{10 classes} & \multirow{2}{*}{\textbf{FD Time}} & \textbf{Runtime} & \textbf{Memory} \\
            & \ mIoU \ & \ mAcc \ & \ mIoU \ & \ mAcc \ & \ mIoU \ & mAcc & & \textbf{(second)} & \textbf{(byte)} \\
            \midrule
            LangSplat & 2.61 & 10.11 & 4.08 & 13.22 & 6.30 & 20.48 & 40 min & 2.1 & 36 \\
            LEGaussians  & 1.62 & 7.26 & 5.72 & 14.23 & 9.84 & 19.13 & 50 min & 1.7 & 32 \\
            OpenGaussian & \sbest 29.43 & 43.62 & 32.61 & 48.26 & 41.29 & 56.42 & 30 min & \sbest 0.003 & \sbest 24 \\
            LUDVIG & 28.47 & 44.17 & 31.47 & 48.54 & 40.47 & 58.15 & 4 min & 0.006 & 2048 \\
            Dr.Splat & 28.00 & \sbest 44.60 & \sbest 38.20 & \sbest 60.40 & \sbest 47.20 & \best 68.90 & \sbest 3 min & - & 128 \\
            Ours & \best 37.11 & \best 58.66 & \best 39.78 & \best 60.91 & \best 47.78 & \sbest 68.21 & \best \textbf{4.1 s} & \best \textbf{0.002} & \best \textbf{2} \\
            \bottomrule
        \end{tabular}
    }
    \label{tab:quantitative_scannet}
\end{table*}

\noindent \textbf{Dataset and Baselines.}
To evaluate 3D scene understanding, we use three benchmarks: LERF-OVS~\cite{qin2024langsplat}, ScanNet~\cite{dai2017scannet}, and DL3DV-OVS, covering both standard and more complex environments.
The first is LERF-OVS, which consists of multi-view images with diverse text query-2D mask annotations for each scene.
The second is ScanNet, a large-scale RGB-D dataset containing 1,500 indoor scenes, each with reconstructed point clouds and per-point semantic labels.
For fair comparison on ScanNet, we follow the evaluation protocol of OpenGaussian and use the same 10 scenes.
For robustness evaluation beyond limited indoor environments, we introduce the DL3DV-OVS dataset.
It is an open-vocabulary extension of DL3DV-10K~\cite{ling2024dl3dv}, which we annotate with 2D mask-text pairs to cover large and complex indoor-outdoor scenes.
The dataset covers a wide range of object scales, distances, and scene complexities across four scenes (park, road, shop, and office), with categories containing varying numbers of instances.

Leveraging these benchmarks, we perform a comprehensive comparison against recent methods, including LangSplat~\cite{qin2024langsplat}, LEGaussians~\cite{shi2024legaussians}, OpenGaussian~\cite{wu2024opengs}, Dr.Splat~\cite{jun2025drsplat}, and LUDVIG~\cite{marrie2025ludvig}.
Since Dr.Splat does not provide inference code, we adopt the reported inference results from its paper and measure all other results ourselves.

\noindent \textbf{3D Understanding Tasks.}
We evaluate performance on two standard tasks.
The first is 3D Object Selection, which aims to segment 3D objects based on open-vocabulary user queries.
For this evaluation, we use the LERF-OVS and DL3DV-OVS dataset, which offer 2D mask-text pairs.
Following previous works~\cite{wu2024opengs, jun2025drsplat}, we identify the 3D Gaussians corresponding to a given text query and render them into 2D.
The rendered results are evaluated using mIoU and mAcc@0.25 against the ground-truth masks.

The second is 3D Semantic Segmentation.
In this task, each 3D Gaussian is classified with the most relevant semantic label, selected from a pool of open-vocabulary text inputs.
Using ScanNet per-point semantic annotations, we evaluate directly in 3D with mIoU and mAcc, and report performance under the 19, 15, and 10 class settings.

\subsection{3D Object Selection}\label{sec:4.2}

\begin{figure*}[!ht]
    \centering
    \includegraphics[width=0.99\linewidth]{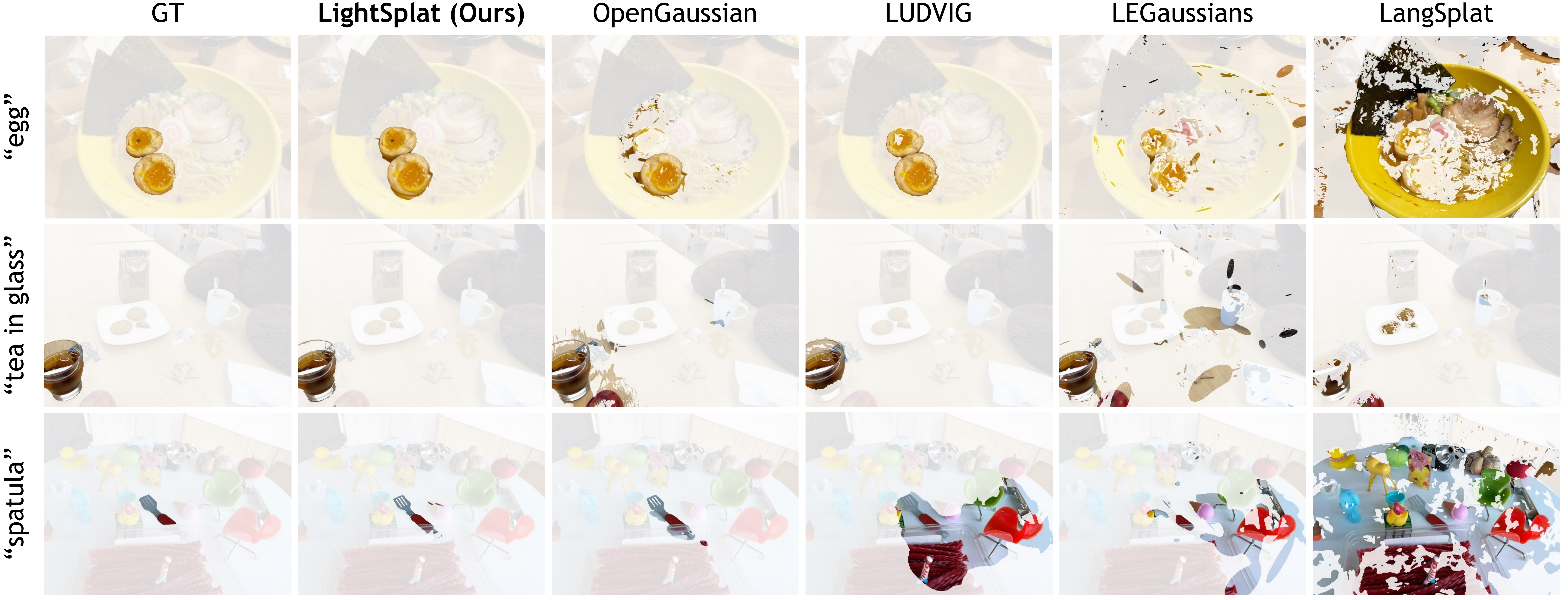}
    \caption{\textbf{Qualitative comparison for 3D Object Selection on the LERF-OVS dataset.} We visualize model performance across different scenes and text queries in LERF-OVS. With context-aware 3D clustering, our method achieves detailed object boundaries while offering significantly faster performance than other methods.
    }
    \label{fig:qualitative_lerf}
\end{figure*}

\begin{figure*}[!ht]
    \centering
    \includegraphics[width=0.99\linewidth]{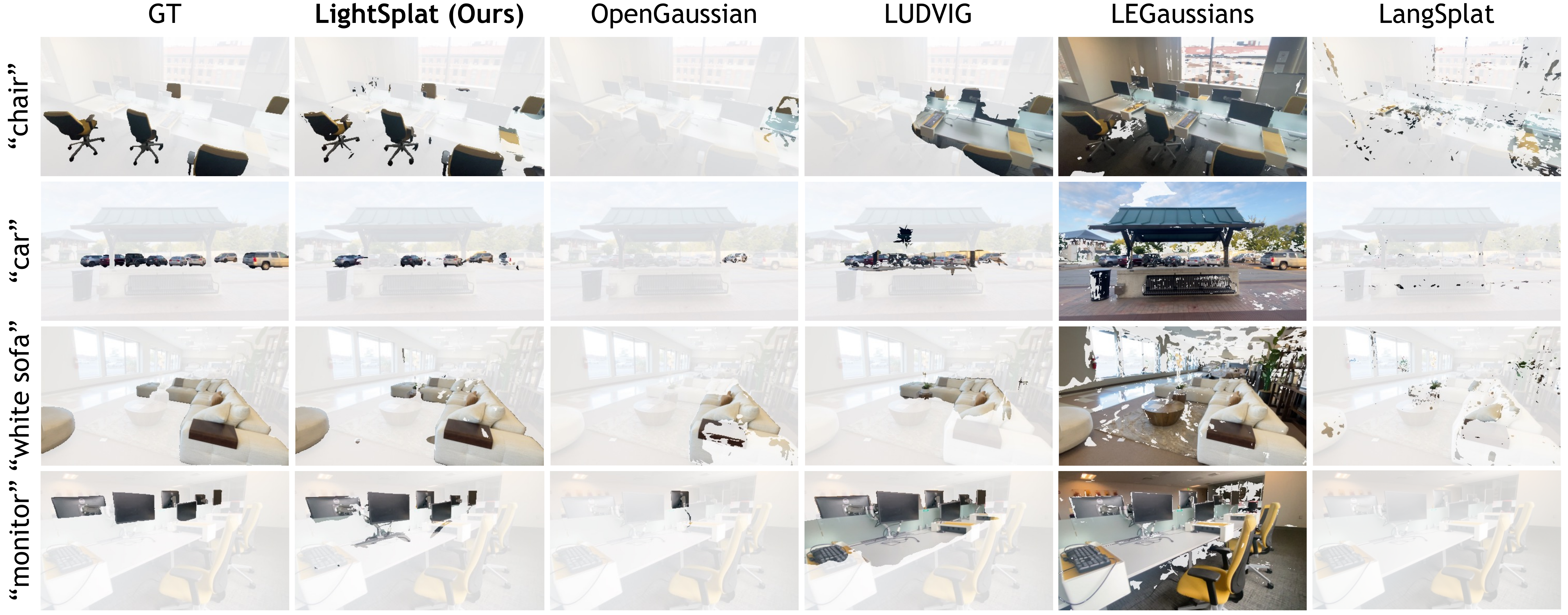}
    \caption{\textbf{Qualitative comparison for 3D Object Selection on the DL3DV-OVS dataset.} 
    We visualize model behavior on large and complex scenes in DL3DV-OVS, covering both indoor and outdoor environments.
    Our method provides reliable selections and clear object boundaries, even in scenes with many similar objects.
    }
    \label{fig:qualitative_dl3dv}
\end{figure*}

\noindent \textbf{Quantitative Results.}
Even without training, our method achieves SOTA performance on LERF-OVS, with a 50$\times$ speedup over recent models, as shown in \Tref{tab:quantitative_lerf}.
In terms of accuracy, we outperform the previous SOTA by an average of 4 mIoU and 4.45 mAcc@0.25.
Notably, our method performs well on challenging scenes like ramen, which contain small and complex objects such as eggs and chopsticks.

As shown in \Tref{tab:quantitative_dl3dv}, our approach also achieves strong performance on DL3DV-OVS, a dataset with large and complex scenes containing multiple objects.
Our method shows robust performance on the road scene with multiple distant cars and the office scene with repeated objects such as chairs and monitors.
Compared to LUDVIG, it achieves 15.77 higher mIoU, 3.93 higher mAcc@0.25, and 150$\times$ faster feature distillation by eliminating iterative feature optimization.
This capacity stems from merging object masks and grouping their associated Gaussians into object-level semantics.
This prevents semantic blurring and preserves clean boundaries even in cluttered or large scenes.
Such results highlight the flexibility and robustness of our method across diverse object scales and scene complexities.

\noindent \textbf{Qualitative Results.}
~\Fref{fig:qualitative_lerf} and ~\Fref{fig:qualitative_dl3dv} present the qualitative comparison on the LERF-OVS and DL3DV-OVS dataset.
Rendering-based methods such as LEGaussians and LangSplat struggle to produce clean segmentations in open-vocabulary scenarios because semantics become blurred when Gaussians are mixed during 2D rendering.
OpenGaussian associates language features in 3D space, but the lack of semantic guidance during clustering leads to inaccurate Gaussian grouping and inconsistent object boundaries.
In contrast, our method produces clean segmentations even for small or complex objects, such as eggs and tea in a glass.
This is achieved by jointly leveraging 2D mask semantics and their corresponding 3D geometry to guide clustering.
Our method remains effective in DL3DV-OVS, maintaining clean boundaries across diverse challenges.
It handles multiple target objects such as chairs and monitors, distant outdoor objects like cars, and large indoor areas with sizable objects like white sofas.

\subsection{3D Semantic Segmentation}\label{sec:4.3}

\begin{figure*}[!ht]
    \centering
    \includegraphics[width=1.0\linewidth]{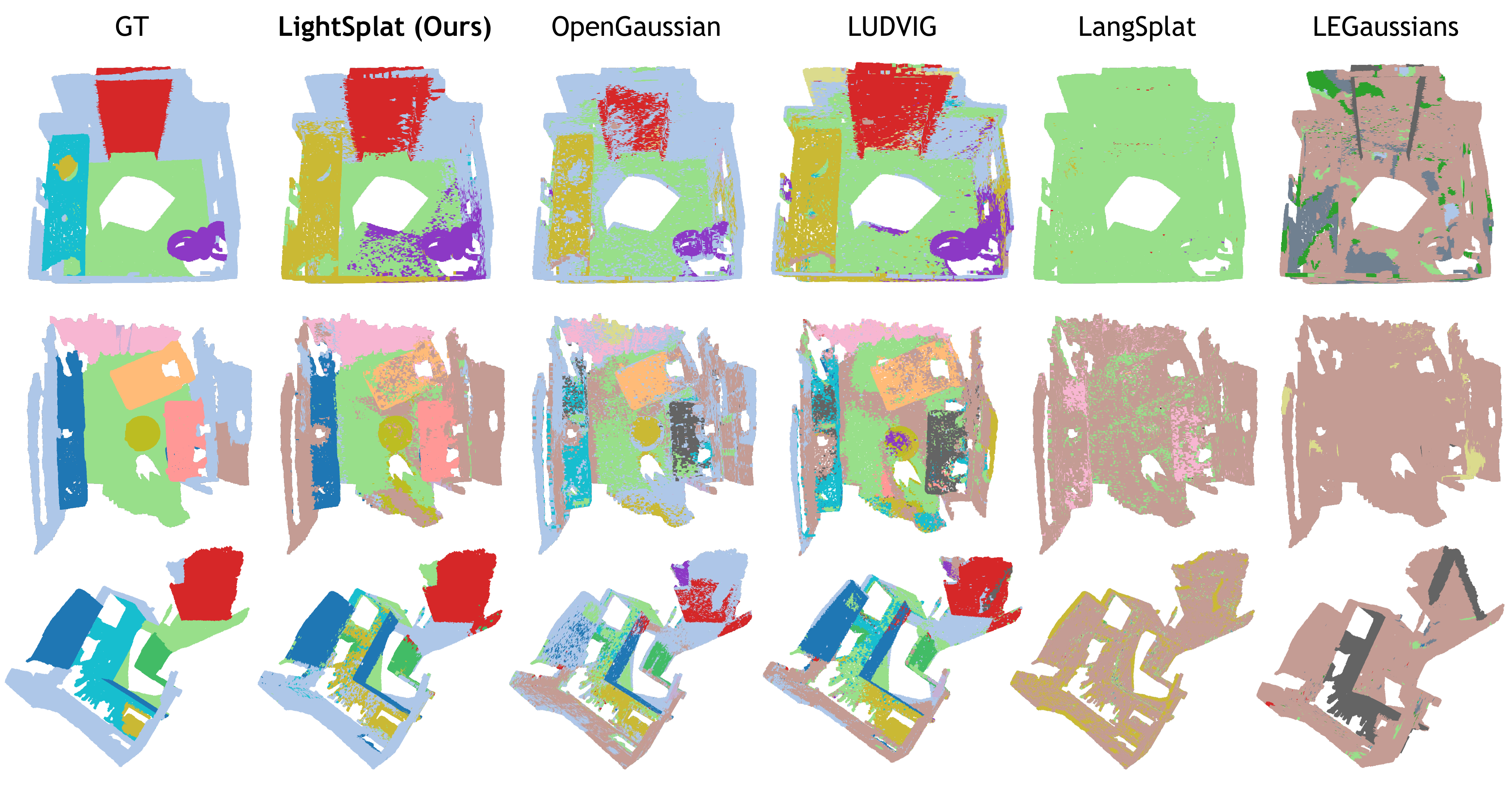}
    \caption{\textbf{Qualitative comparison for 3D Semantic Segmentation on the ScanNet dataset.} For intuitive comparison, ground-truth semantics are visualized in distinct colors, and predictions from each model are shown side by side.
    Compared to other methods, our approach more effectively captures both object (e.g., door) and large-area semantics (e.g., wall), demonstrating robustness across diverse real-world scenes and adaptability to a wide range of queries.
    }
    \label{fig:qualitative_scannet}
\end{figure*}

\noindent \textbf{Quantitative Results.}
~\Tref{tab:quantitative_scannet} presents the quantitative results on the ScanNet dataset.
Our method achieves the best performance across all settings.
In the most challenging 19-class configuration, it surpasses the previous SOTA by 7.68 mIoU and 14.06 mAcc, highlighting its strong 3D segmentation capability.
In addition, our approach reduces feature distillation to only 4.1 seconds and improves memory efficiency by up to 64$\times$, while supporting fast inference at 500 FPS for text-query retrieval using precomputed cluster features.
These advantages stem from avoiding iterative per-Gaussian feature aggregation and high-dimensional feature handling, which are major bottlenecks in LUDVIG and Dr.Splat.
Instead, our index-based design groups Gaussians into object-level units using compact semantic indices, enabling rapid feature distillation and real-time inference.
These results further show that our method remains robust across text queries from object-centric descriptions to indoor spatial semantics, delivering fast and scalable performance for practical 3D applications.

\noindent \textbf{Qualitative Results.}
~\Fref{fig:qualitative_scannet} presents the qualitative results on the ScanNet dataset.
Rendering-based methods like LEGaussians and LangSplat identify only a few object classes, as 2D semantic distillation struggles to separate objects in 3D space.
OpenGaussian and LUDVIG better distinguish object regions but still suffer from blurry boundaries and mislabeling.
Since these methods use semantics at the level of individual Gaussians, they fail to form meaningful object structure, making object-level reasoning difficult.
In contrast, our method captures semantics more consistently, handling both object-level categories (e.g. doors) and large spatial regions (e.g. walls and floors).

\subsection{Ablation Study}\label{sec:4.4}

\begin{table}[t]
    \centering
    \caption{\textbf{Ablation Study on LERF-OVS dataset.} The results demonstrate that each component not only boosts performance, but also directly contributes to the fast FD time of our method.
    FD Time refers to feature distillation time.
    }
    \resizebox{0.90\linewidth}{!}{ 
        \begin{tabular}{c|ccc}
            \toprule
            Ablation & mIoU & Acc@0.25 & FD Time \\
            \midrule
            Ours Full & \best 47.58 & \best 68.32 & 4.55 s \\
            w/o 3D Mask Filtering & \sbest 29.31 & \sbest 25.96 & 4.54 s \\
            w/o Semantic-Aware & 19.56 & 18.97 & \best 4.42 s \\
            w/o Geometry-Aware & 2.01 & 2.27 & \sbest 4.44 s \\
            \bottomrule
        \end{tabular}
    }
    \label{tab:ablation}
\end{table}

We conduct an ablation study by removing each component individually.
As shown in \Tref{tab:ablation}, each component significantly contributes to the overall performance.
Without 3D-aware mask filtering, noisy masks degrade clustering quality, disrupting 2D-3D alignment and reducing mIoU from 47.58 to 29.31.
Removing semantic-aware clustering decreases performance by over 50\%, as the model cannot identify semantically corresponding masks across views for merging.
Without geometry-aware clustering, the model ignores mask overlap in 3D, which prevents it from leveraging 2D-3D correspondences or maintain spatial consistency.
As a result, performance drops sharply to 2.01 mIoU.

Notably, the feature distillation time is almost the same across all variants, differing by only about 0.1 seconds.
This reflects the efficiency and necessity of our design.
Most of the computation lies in evaluating Gaussian contributions to the 2D masks, which is inherently required, while the remaining steps add only negligible overhead.

%% file: sec/5_conclusion.tex
\section{Conclusion}

We have proposed \texttt{LightSplat}, a fast and memory-efficient training-free framework for open-vocabulary 3D scene understanding.
By directly injecting compact 2-byte mask indices into 3D Gaussians based on their influence, and managing semantics through an index-feature mapping, our method eliminates the need for iterative optimization and high-dimensional per-Gaussian storage.
Furthermore, our context-aware clustering uses geometric and semantic cues to form cluster-level semantics from per-mask features, enabling fast and interpretable inference.
Extensive experiments demonstrate that \texttt{LightSplat} achieves state-of-the-art performance with 50-400$\times$ faster feature distillation and 64$\times$ lower memory usage, offering a scalable and practical foundation for real-time 3D applications.

%% file: sec/6_ack.tex
\section*{Acknowledgements}\label{sec:ack}

This work was supported by Institute of Information \& communications Technology Planning \& Evaluation (IITP) grant funded by the Korea government (MSIT) (No.RS-2020-II201336, Artificial Intelligence Graduate School Program (UNIST); 
No.RS-2025-25442824, AI Star Fellowship Program (UNIST);
No.RS-2025-25442149, LG AI STAR Talent Development Program for Leading Large-Scale Generative AI Models in the Physical AI Domain), and by the InnoCORE program of the Ministry of Science and ICT (N10250156).

%% file: sec_supp/1_dataset.tex
\section*{Overview}
In these supplementary materials, we provide additional details on the following topics:
\begin{itemize}
    \item \Sref{sec:dataset} introduces the DL3DV-OVS dataset, a large-scale open-vocabulary benchmark covering four indoor-outdoor scenes with diverse object queries.
    \item \Sref{sec:imple_detail} outlines our experimental setup, including 3DGS, feature extraction, and hyperparameters.
    \item \Sref{sec:more_quan} summarizes feature distillation time and reports millisecond-level inference from lightweight clustering.
    \item \Sref{sec:more_qual} presents additional qualitative comparisons that demonstrate robustness in diverse 3D scenes.
    \item \Sref{sec:edit} showcases text-driven 3D editing enabled by cluster-level semantic control.
    \item \Sref{sec:limitation} discusses limitations regarding object feature selection and the precision-speed trade-off.
\end{itemize}

\section{DL3DV-OVS Dataset}\label{sec:dataset}

We propose DL3DV-OVS, an open-vocabulary extension of DL3DV for evaluating model robustness in large-scale and challenging scenes.
The dataset spans indoor and outdoor environments with broad spatial layouts and diverse scene structures.
It provides 22 text queries for objects such as cars, chairs, monitors, and streetlights.
These items vary significantly in size and distance, often appearing multiple times within a single view.
DL3DV-OVS consists of four scenes: park, road, office, and shop.
For each scene, we provide 960$\times$540 multi-view images, text queries, and the corresponding ground-truth 2D masks.
Table~\ref{tab:dl3dv_queries} lists the text queries for each scene.

\begin{table}[!ht]
    \centering
    \caption{\textbf{Scene-wise text queries in the DL3DV-OVS dataset.}
    Each scene provides a set of text queries for evaluating open-vocabulary 3D object selection.
    }
    \resizebox{1.0\linewidth}{!}{ 
        \begin{tabular}{c|lll}
            \toprule
            \textbf{Scene} & \multicolumn{3}{c}{\textbf{Text Queries}} \\
            \midrule
            
            \multirow{2}{*}{Park} &
            bench & trash bin & car \\
            & parking signpost & information board \\
            \midrule
            
            \multirow{3}{*}{Shop} &
            black drawer & framed artwork & lamp \\
            & potted plant & white sofa & bed \\
            & fire extinguisher & \multicolumn{2}{l}{round wooden coffee table} \\
            \midrule
            
            \multirow{3}{*}{Road} &
            bench & car & fire hydrant \\
            & planter & roof & trash bin \\
            & streetlight \\
            \midrule
            
            \multirow{2}{*}{Office} &
            chair & keyboard & monitor \\
            & whiteboard & mouse \\
            \midrule
        \end{tabular}
    }
    \label{tab:dl3dv_queries}
\end{table}

%% file: sec_supp/2_implementation.tex
\section{More Implementation Details}\label{sec:imple_detail}

\subsection{Experimental Setup}

\noindent \textbf{Training Setup.} \ 
For consistent comparison of speed and accuracy, all models are evaluated on a single RTX 4090 GPU.
LUDVIG is the only exception, running on an A6000 because its high-dimensional language feature operations exceed the 24GB memory limit.

\vspace{1mm} \noindent \textbf{3D Gaussian Splatting.} \ 
To ensure a fair comparison, we use a pre-trained 3DGS trained for 30,000 iterations, identical to LangSplat and OpenGaussian.
We do not modify any Gaussian parameters and simply allocate an additional 2-byte space per Gaussian to store an index.
A full language feature requires 2028 bytes per Gaussian, so storing it for 100,000 Gaussians would require about 200 MB, which is much larger than the 23.6 MB needed for all standard Gaussian parameters.
In our method, we no longer store this 2024-byte feature for each Gaussian.
Instead, each Gaussian stores only a small index, with the total index size being just 0.2 MB, which is less than 0.1\% of the 200 MB required for per-Gaussian language features.

\vspace{1mm} \noindent \textbf{Object Feature Extraction.} \ 
LangSplat uses all three hierarchical mask levels produced by SAM and learns separate language features for each level.
To ensure both fair comparison and computational efficiency, we follow OpenGaussian and adopt a large-mask strategy.
We extract mask-level CLIP features by cropping each SAM mask, resizing it to $224\times224$ with zero padding, and encoding it using OpenCLIP~\cite{cherti2023openclip}.
With this setup, our method integrates semantics and geometry without relying on such hierarchical structures, achieving SOTA performance with 50$\times$ faster speed.

\subsection{Hyperparameters}
To ensure fair evaluation, we use the same hyperparameters for all scenes within each dataset.
Denser views or more compact scenes produce higher mask overlap and thus require stricter IoU thresholds.
In contrast, datasets with more diverse text queries require slightly relaxed feature similarity thresholds.
Accordingly, for LERF-OVS, ScanNet, and DL3DV-OVS, we set (contrib, noise, IoU, feat) to (0.04, 200, 0.6, 0.75), (0.04, 500, 0.35, 0.8), and (0.09, 450, 0.5, 0.8), respectively.
Unlike prior methods that tune hyperparameters per scene (e.g., teatime)~\cite{wu2024opengs}, our method uses fixed settings and still performs consistently well across diverse scenes, including the challenging ramen scene.
Across all three datasets, our method achieves a 400-720$\times$ speedup in feature distillation and improves mIoU over OpenGaussian by 5.43, 7.68, and 19.65 points, respectively.
These results demonstrate strong robustness across diverse environments.

\subsection{Evaluation Setup}
\noindent \textbf{Model Comparison.} \ 
Since Dr.Splat provides only training code, we adopt the reported inference metrics from its paper and measure all other results ourselves.
Dr.Splat is expected to have higher inference time, as it processes 128-dimensional features through a codebook and compares them against all Gaussians.
In contrast, our method performs semantic comparison at the 3D cluster level instead of evaluating every individual Gaussian.
By attaching semantics to clusters through an index-feature mapping without relying on a codebook, the comparison set remains compact and efficient.
As shown in Table 3, this enables our method to achieve a 2ms inference time on average across ScanNet scenes.

\vspace{1mm} \noindent \textbf{3D Object Selection.} \ 
This section provides a detailed explanation of the 3D Object Selection task.
Previous works, such as LangSplat and LEGaussians, struggle to accurately distinguish objects in 3D space.
This is because these methods distill features after rendering 3D Gaussians into 2D, where overlapping Gaussians blur the features and result in indirect, less effective supervision.
As a result, high-quality features extracted in 2D often do not transfer well to 3D, and 2D-based evaluation provides only a partial view of the model’s 3D understanding.
To address this, we adopt the 3D Object Selection task, which directly selects objects in 3D from text queries.
We evaluate on LERF-OVS by first selecting objects directly in 3D, then rendering them back to 2D for comparison with ground-truth masks, following OpenGaussian and Dr.Splat.

\begin{figure}[t]
    \centering
    \includegraphics[width=1.0\linewidth]{fig_supp/1_module_FD_time.png}
    \caption{\textbf{Module-wise feature distillation time.}
    This figure shows the time proportion of each module during feature distillation.
    Injection, filtering, and clustering refer to indexed feature injection, 3D-aware mask filtering, and context-aware 3D clustering, respectively.
    }
    \label{fig:module_FD_time}
\end{figure}

\begin{table}[!ht]
    \centering
    \caption{\textbf{Inference time comparison on LERF-OVS and DL3DV-OVS.}
    We report inference time in seconds.
    While prior methods rely on slow 2D or per-Gaussian processing, \texttt{LightSplat}'s cluster-level inference achieves millisecond-level runtime.
    }
    \resizebox{0.80\linewidth}{!}{ 
        \begin{tabular}{c|cc}
            \toprule
            \textbf{Methods} & \textbf{LERF-OVS} & \textbf{DL3DV-OVS} \\
            \midrule
            LangSplat & 1.046 & 0.175 \\
            LEGaussians & 0.383 & 0.770 \\
            LUDVIG & 5.923 & 1.865 \\
            OpenGaussian & \sbest 0.003 & \sbest 0.005 \\
            Ours & \best 0.001 & \best 0.003 \\
            \bottomrule
        \end{tabular}
    }
    \label{tab:inference_time}
\end{table}

\vspace{1mm} \noindent \textbf{3D Semantic Segmentation.} \ 
We evaluate 3D semantic segmentation on the ScanNet benchmark, following the evaluation protocol used in OpenGaussian.
The protocol fits 3D Gaussians directly to the ScanNet ground-truth points, matching their locations and counts.
This establishes a one-to-one correspondence between Gaussians and GT points, enabling direct comparison of predicted and ground-truth labels.
For fair comparison, all baselines follow the same evaluation setting.

%% file: sec_supp/3_quantitative.tex
\section{More Quantitative Results}\label{sec:more_quan}

\noindent \textbf{Module FD Time.} \ 
Fig.~\ref{fig:module_FD_time} shows the average time consumed by each module in our feature distillation pipeline across the LERF-OVS scenes.
Most of the computational cost arises from rendering and indexed feature injection, since both steps require computation for each view.
In contrast, the subsequent 3D-aware mask filtering and context-aware clustering operate in a single step, adding almost no additional cost or delay.
This design keeps the overall distillation process extremely fast, and the ablation study shows that these components improve performance without slowing the pipeline.
Across scenes, rendering, injection, filtering, and clustering account for 36.5\%, 53.7\%, 3.8\%, and 6.0\% of the total time, respectively.

\vspace{1mm} \noindent \textbf{Inference Time.} \ 
Table~\ref{tab:inference_time} presents inference-time results on LERF-OVS and DL3DV-OVS, providing additional evaluations beyond the ScanNet results included in the main paper.
Rendering-based methods such as LangSplat and LEGaussians incur significant computational overhead because their inference pipeline requires additional decoding of semantic features.
LangSplat performs per-Gaussian latent decoding through its autoencoder, while LEGaussians applies per-pixel semantic decoding using its MLP, both of which further increase latency.
OpenGaussian is also slow because it computes high-dimensional features for every Gaussian.
LUDVIG introduces additional overhead by diffusing 2048-dimensional features over a graph.
In contrast, \texttt{LightSplat} performs cluster-level inference through an index-feature mapping, avoiding per-Gaussian comparisons and diffusion.
As a result, \texttt{LightSplat} maintains millisecond-level inference even in large indoor-outdoor scenes.

%% file: sec_supp/4_qualitative.tex
\section{More Qualitative Results}\label{sec:more_qual}

We present additional qualitative results omitted from the main paper due to space constraints.
These results demonstrate the robustness of our model across a wide range of scenes.

\vspace{1mm}  \noindent \textbf{Clustering Process.}
Fig.~\ref{fig:mask2cluster} illustrates how \texttt{LightSplat} converts 2D masks into stable 3D object clusters.
We use the same hyperparameters for the two DL3DV-OVS scenes shown above and the two additional indoor-outdoor scenes below, demonstrating robustness without scene-specific tuning.
The process begins with SAM masks extracted from the input images.
These masks are injected into the 3D scene based on per-pixel contribution, producing a mask ID field where each Gaussian receives the index of its most influential 2D mask.
This intermediate field may appear noisy due to view-dependent inconsistencies.
We address this by filtering out unstable masks in the 3D-aware mask filtering stage using 2D-3D correspondences.
Finally, context-aware 3D clustering groups semantically and spatially consistent masks into clean object-level clusters.
The resulting cluster ID field demonstrates robust and coherent semantics across diverse indoor and outdoor environments.

\vspace{1mm}  \noindent \textbf{3D Object Selection.} \ 
Fig.~\ref{fig:more_qualitative_lerf} and Fig.~\ref{fig:more_qualitative_dl3dv} present additional qualitative results on the LERF-OVS and DL3DV-OVS datasets.
Unlike models where iterative aggregation blurs semantics in 3D~\cite{qin2024langsplat, shi2024legaussians, bhalgat2024n2f2, jun2025drsplat, marrie2025ludvig}, our method transfers mask-level semantics directly into 3D.
It then merges related masks into object-level clusters, preserving much clearer object structure than Gaussian-level aggregation.
As a result, our method achieves high accuracy in challenging real-world cases:
\begin{itemize}
    \item small objects attached to others (kamaboko, wavy noodle)
    \item multiple instances of the same object (knives, cars)
    \item semantically subtle categories (yellow pouf)
    \item thin or complex structures (jake with thin legs, streetlight)
    \item single outdoor objects (trash bin, information board)
\end{itemize}

\vspace{1mm}  \noindent \textbf{3D Semantic Segmentation.} \ 
\Fref{fig:more_qualitative_scannet} presents more qualitative results on the ScanNet dataset.
These results show that our approach captures a broader range of semantics in indoor scenes, while also producing more complete object coverage with well-aligned boundaries.

These additional qualitative results further highlight the robustness of our model to diverse and challenging queries, from small objects to complex structures, while outperforming other models in both accuracy and speed on LERF-OVS, ScanNet, and DL3DV-OVS.

%% file: sec_supp/5_applications.tex
\section{3D Scene Editing}\label{sec:edit}

\Fref{fig:3d_editing} shows how our method naturally extends to 3D scene editing. 
By managing semantics at the cluster level, our method enables fast and accurate segmentation of text-specified objects in 3D space.
Building on this segmentation capability, users can directly remove, recolor, or reposition the identified objects, enabling high-fidelity scene edits.
We demonstrate this capability through recoloring and enlarging as representative examples.
Enlargement is performed by adjusting Gaussian distances and scales, while recoloring is achieved by modifying SH coefficients.
Notably, these editing operations add minimal overhead, keeping inference time nearly unchanged in interactive scenarios.
Together, these results highlight the speed, accuracy, and flexibility of our approach for interactive 3D manipulation in immersive content creation, AR/VR, and robotics.

%% file: sec_supp/6_limitations.tex
\section{Limitations}\label{sec:limitation}

\noindent \textbf{Object Feature Selection.} \ 
SAM provides high-quality object segmentation masks, and CLIP enables effective image-text interaction as a foundation model trained on large-scale datasets in a shared embedding space.
However, SAM masks do not always produce perfectly accurate boundaries.
In addition, CLIP can struggle to distinguish between semantically similar objects (e.g., red apple vs. green apple) due to their high feature similarity.
These limitations could be further mitigated with extra modules or iterative training, but this would disrupt the speed–accuracy balance of our approach.

\vspace{1mm} \noindent \textbf{Hyperparameters.} \
Our method performs semantic injection and 3D clustering in a single step using only four main hyperparameters, making it lightweight and intuitive.
Compared to training-based methods, which involve numerous hyperparameters for model initialization and optimization, our approach requires far fewer parameters, avoiding the high training cost and memory usage.
Despite its simplicity, our method can still be influenced by the choice of hyperparameters and can be further refined for improvement.